\pdfoutput=1
\documentclass[11pt]{article}

\usepackage{acl}

\usepackage{times}
\usepackage{latexsym}

\usepackage[T1]{fontenc}
\usepackage[utf8]{inputenc}

\usepackage{microtype}
\usepackage{graphicx}
\usepackage{multirow,array}
\usepackage{bbm}
\usepackage{algorithm,algorithmicx}
\usepackage{amsmath}
\usepackage{amsthm}
\usepackage{verbatim}
\usepackage{booktabs}
\usepackage{amsfonts,amssymb} 
\usepackage[noend]{algpseudocode}
\usepackage{bm}
\usepackage{enumitem}

\newcommand{\secref}[1]{Section \ref{#1}}
\newcommand{\figref}[1]{Figure \ref{#1}}

\newcommand{\tabref}[1]{Table \ref{#1}}

\title{On the Efficacy of Eviction Policy for Key-Value Constrained Generative Language Model Inference}

\author{Siyu Ren \\
Shanghai Jiao Tong University, China\\
\texttt{roy0702@sjtu.edu.cn} \\
\And
Kenny Q. Zhu\textsuperscript{\rm}\thanks{\hspace{2mm}The corresponding author.}\\
University of Texas at Arlington, USA\\
\texttt{kenny.zhu@uta.edu} \\
}

\begin{document}
\maketitle
\begin{abstract}
  Large language models~(LLMs) are notably cost-prohibitive to deploy in resource-constrained environments due to their excessive memory and computational demands. In addition to model parameters, the key-value cache is also stored in GPU memory, growing linearly with batch size and sequence length. As a remedy, recent works have proposed various eviction policies for maintaining the overhead of key-value cache under a given budget. 
 This paper embarks on the efficacy of existing eviction policies in terms of \textit{importance score calculation} and \textit{eviction scope construction}. We identify the deficiency of prior policies in these two aspects and introduce RoCo, a \underline{r}\underline{o}bust \underline{c}ache \underline{o}mission policy based on temporal attention scores and robustness measures. Extensive experimentation spanning prefilling and auto-regressive decoding stages validates the superiority of RoCo. Finally, we release EasyKV, a versatile software package dedicated to user-friendly key-value constrained generative inference. Code available at \url{https://github.com/DRSY/EasyKV}.
\end{abstract}

\section{Introduction}
Recent advancements in Large Language Models~(LLMs) have demonstrated outstanding proficiency in a wide range of text generation scenarios~\cite{thoppilan2022lamda,wei2022emergent,wang-etal-2023-self-instruct,sun2023corex,wu2024copilot}. Nevertheless, deploying LLMs is a notably costly undertaking considering their tremendous parameter size and quadratic cost of attention layers. Accordingly, model compression~\cite{frantar2023massive,xia2023sheared} and memory-efficient attention~\cite{dao2022flashattention,dao2023flashattention} techniques have emerged to tackle these challenges and achieved substantial outcomes.

\begin{figure}[t]
    \centering
    \scalebox{0.5}{\includegraphics{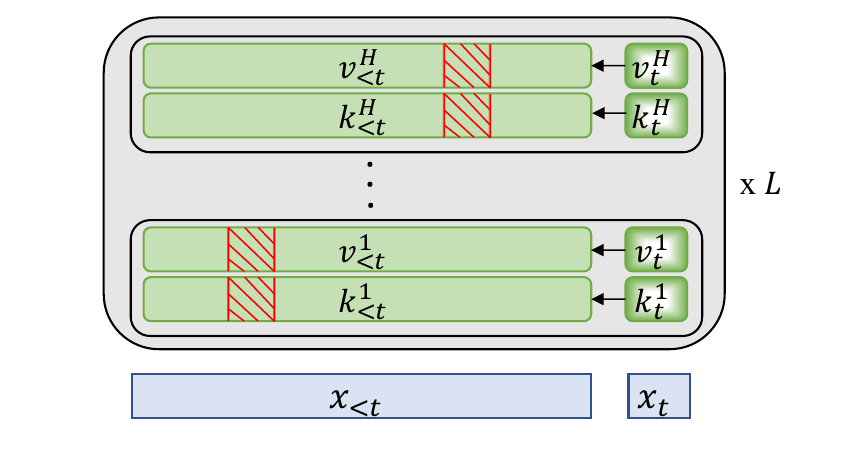}}
    \caption{Illustration of KV cache eviction inside one attention layer~($L$ in total). In this example, a single pair of key-value vectors are deleted~(red hatched areas) before appending the next token's. Different heads~($H$ in total) at model layers may evict at different positions.}
	\label{fig:intro}
\end{figure}
Owning to the auto-regressive nature of LLM inference~\cite{vaswani2017attention,radford2019language}, the intermediate attention key-value vectors are also required to be stored in memory to avoid redundant key-value projection in future steps. The size of key-value cache~(KV cache) depends on the configuration of the attention layer, batch size, and sequence length, which poses challenges in both memory footprint, I/O cost, and computation burden given the increasingly upsoaring model scale and user requests. Variants like multi-query attention and grouped-query attention~\cite{mqa,gqa} reduce the size of the KV cache with fewer attention heads, but cannot be directly applied to pre-trained LLMs without re-training. 

In pursuit of flexible and training-free control over KV cache during LLM inference, recent works~\cite{liu2023scissorhands,h2o,xiao2023efficient,tova} have investigated implementing KV cache with \textit{eviction policy}, where the key-value vectors of certain tokens are strategically deleted from memory~(see \figref{fig:intro}). In this way, the size of the KV cache can be maintained under a specified budget, leading to decreased memory and computational overhead. Despite the claimed and empirically observed reduction in KV cache, there still lacks a comprehensive comparative analysis of these methods. This study aims to fill this gap and embarks on the efficacy of existing eviction policies from a unified framework, which decomposes an eviction policy into two design dimensions: \textit{importance score calculation} and \textit{eviction scope construction}. 
The former characterizes how important a pair of key-value vectors is to future generations, 
while the latter determines which tokens are readily allowed to be evicted from the cache
We categorize existing eviction policies according to these two dimensions. 
Following our preliminary analysis, we discover that the way current methods calculate 
importance scores utilizing local statistics can only weakly approximate that derived 
from global statistics~(full KV cache without eviction). 
Moreover, prior methods commonly 
construct the eviction scope by only incorporating tokens outside of a local window, 
which we show endure high sensitivity to window size. 

In this paper, we propose RoCo, a \underline{R}\underline{o}bust \underline{C}ache \underline{o}mission policy based on local attention scores and robustness measures. Specifically, we compute the importance score of each in-cache token using averaged attention probability from future tokens, 
and formulate the eviction scope using tokens with the lowest variance of attention. 
RoCo exhibits significantly better consistency with 
full KV cache counterpart and robustness to eviction scope size.

To evaluate the effectiveness of RoCo in terms 
of preserving the LLM's downstream performance, we perform experiments across 
both prefilling and auto-regressive decoding stages at which cache eviction happens, 
spanning tasks of language modeling, text summarization, context reconstruction, and 
instruction following. Experimental results at different levels of KV cache 
budget demonstrates that RoCo results in significantly better generation 
quality compared to current methods judged by both automatic metrics and LLM-based evaluator.

Our contributions are summarized as follows:
\begin{itemize}[itemsep=1pt,parsep=2pt,topsep=1pt]
    \item We systematically analyze current cache eviction policies from the dimensions of importance score calculation and eviction scope construction, shedding light on their limitations.
    \item Based on our analysis, we introduce a robust cache omission policy named RoCo and conduct a comprehensive evaluation to verify its effectiveness on downstream tasks.
    \item We open-source EasyKV, a versatile software package that supports key-value constrained generative LLM inference with flexible configuration on cache budget and eviction policy.
\end{itemize}
\section{Background}
In this section, we present necessary background about Transformer as well as existing literature on addressing the memory and computational bottleneck of Transformer-based LLMs.

\subsection{Transformer-based LLMs}
The input to a Transformer-based LLM is a sequence of tokens 
$\bm{x}=(x_1,...,x_{T})$, which is further processed by the embedding layer, followed by a series of Transformer blocks composed of an attention block and a feedforward block. The attention block is the only submodule where tokens at different positions exchange information, necessitating the need for a key-value cache during inference.
\paragraph{Attention Block}
At the $l$-th layer, the input hidden states $\bm{H}^{l-1}\in \mathbb{R}^{T\times d}$ is multiplied with three matrices $\bm{W}_{q}^{l}, \bm{W}_{k}^{l}$, and $\bm{W}_{v}^{l}$, producing $\bm{Q}^{l}=\bm{H}^{l-1}\bm{W}_q^l, \bm{K}^{l}=\bm{H}^{l-1}\bm{W}_k^l, \bm{V}^{l}=\bm{H}^{l-1}\bm{W}_v^l$. Then the scaled dot-product attention is performed as follows:
\begin{align}
    \text{Attn}_i&=\text{Softmax}(\frac{\bm{Q}^{l}_{i}\cdot (\bm{K}_{i}^{l})^{\top}}{\sqrt{d^\prime}})\cdot \bm{V}_{i}^{l} \\
    \text{SDPA}&=\text{Concat}(\text{Attn}_1,...,\text{Attn}_{H})\cdot \bm{W}_{o}^{l}
\end{align}
where $H$ is the number of attention heads, $d^\prime=\frac{d}{H}$ is the head dimension, and $\bm{W}_o^l$ is the output matrix.
\paragraph{Key-Value Cache}
LLM inference follows an autoregressive fashion. During training, it masks the upper triangular part of the attention matrix such that each token only sees itself and previous tokens. At inference time, 
the common practice is to cache the key-value vectors computed so far and append the newly computed ones into the cache. At time step $T$, the key-value cache can be written as a tensor of shape $(L, 2, B, H, T, d^\prime)$, where $L$ is the number of model layers and $B$ is the batch size. 
It is evident that the size of the KV cache grows linearly with respect to sequence length, potentially leading to excessive memory and latency issues when dealing with long input or output.

\subsection{Efficient LLMs}
Recent years have witnessed a surge of studies attempting to optimize the inference cost of LLMs from different~(often orthogonal) perspectives.

One line of work follows the conventional model compression paradigm, aiming to identify and remove redundancy from billions of model parameters. These include tensor decomposition~\cite{dao2022monarch}, weight pruning~\cite{frantar2023massive,xia2023sheared,slicegpt}, and quantization~\cite{dettmers2022llm,gptq,smoothquant}. These methods reduce the KV cache footprint by reducing the model dimension, layers, and data precision.

Another line of work focuses on architectural design, aiming at reducing model complexity from the ground up. Representatives include sparse attention Transformers~\cite{child2019generating,bigbird}, linear attention Transformers~\cite{linformer,performer,qin-etal-2022-devil}, and simplified attention variants~\cite{mqa,gqa}. These methods either completely eschew the $O(T)$ space complexity of KV cache size or reduce the number of attention heads in exchange for a larger context length.

Some recent efforts~\cite{liu2023scissorhands,h2o,tova} pay attention to methods that maintain the memory usage of the KV cache under a fixed budget without finetuning or architectural modifications to the model. The shared tenet of these approaches is the discernment and retention of key-value vectors that exert a significant influence on future generations.
This work follows this line of research, dissects the efficacy of existing eviction policies, and introduces an improved policy with more consistent importance score and robust eviction scope construction.

\section{Problem Formulation}
\paragraph{Standard Inference} Denote the input prompt to the LLM as $\bm{x}=(x_1,x_2,...,x_{T})$, the standard generative inference process consists of two consecutive stages: prefilling and auto-regressive decoding. The prefilling stage encodes the input prompt $\bm{x}$ and produces the corresponding attention key matrix $\bm{K}_T\in \mathbb{R}^{L\times H\times T\times d^\prime}$ and value matrix $\bm{V}_T\in \mathbb{R}^{L\times H\times T\times d^\prime}$, 
where $L$, $H$, $d^\prime$ represent the number of model layers, number of attention heads, and per-head dimension, respectively. 
Afterward, the LLM samples one token from its output distribution at each step conditioned on all key-value states computed so far. The key-value matrices are updated by appending the key-value vectors of this new token:
\begin{align}
    x_{T+1}&\sim \text{LLM}(\cdot | x_{<=T}) \\
    \bm{K}_{T+1}&=\text{Concat}(\bm{K}_T, K_{T+1}) \\ 
    \bm{V}_{T+1}&=\text{Concat}(\bm{V}_T, V_{T+1})
\end{align}
where $K_{T+1}\in \mathbb{R}^{L\times H\times 1\times d^\prime}$, $V_{T+1}\in \mathbb{R}^{L\times H\times 1\times d^\prime}$ are the key and value vectors of $x_{T+1}$. 
The above process is repeated until the end of sequence token is generated. Let $\tilde{\bm{x}}=(\bm{x}, \bm{x}_o)$ denote the complete token sequence composed of input prompt $\bm{x}$ and output $\bm{x}_o$, where the output sequence contains $N$ tokens. 
The peak cache size during standard inference is therefore determined by the key-value matrix $\{\bm{K}_{T+N},\bm{V}_{T+N}\}\in \mathbb{R}^{L\times H\times (T+N)\times d^\prime}$.

\paragraph{Key-Value Constrained Inference}
LLMs are typically deployed on hardware with constrained memory resources. However, during standard generative inference, the size of the key-value cache increases linearly with the total length of the sequence, potentially leading to out-of-memory issues and the associated latency incurred by reading and writing between High Bandwidth Memory (HBM) and Static Random Access Memory (SRAM)~\cite{dao2022flashattention}.

To this end, recent studies have shifted toward key-value-constrained inference as a more controllable inference scheme. Denoting the fixed budget for each attention head as $B$ tokens, key-value constrained inference is to maintain the key-value matrices $\bm{K}_t$ and $\bm{V}_t$ such that 
$\bm{K}_t$, $\bm{V}_t\in \mathbb{R}^{L\times H\times n\times d^\prime}$ and $n\leq B$ for any $t\in\{1,...,T+N\}$. 

\section{Eviction Policy for Key-Value Constrained Inference}
In practice, $\bm{K}_t$ and $\bm{V}_t$ are stored in a fixed memory buffer with a maximum token budget $B$. When the buffer is full, an eviction policy is executed to remove stored but non-influential tokens from the cache. Although various eviction policies have been proposed, there still lacks a systematic comparison of their working mechanisms, design choices, and downstream performance. 

To fill this gap, we embark on the efficacy of existing eviction policies from a unified 
framework. Concretely, we represent an eviction policy as the composition of two components: 
importance score calculation and eviction scope construction, which we elaborate 
on in the following sections. 
\subsection{Importance Score Calculation}
\label{sec:isc}
Importance score calculation plays a vital role in eviction policy. It determines the relative order by which tokens are evicted. We summarize existing importance score calculation methods as follows: \\
\textbf{Random Deletion}~~As a naive baseline, one can randomly choose the key-value vectors to evict. We incorporate this method into comparison and let it serve as the lower bound. \\
\textbf{Recency}~~This method deems the farthest token as least important and evicts it when the buffer is full. It is also referred to as window attention in prior studies~\cite{etc,longformer,xiao2023efficient}. \\
\textbf{Accumlative Attention Score~(AAS)}~~$\text{H}_{\text{2}}$O~\cite{h2o} maintains a $B$-sized record array that stores the accumulative attention score each in-cache token received from subsequent tokens. \\
\textbf{Accumlative Quantized Attention Score~(AQAS)}~~ScissorHands~\cite{liu2023scissorhands} adopts an apporach similar to $\text{H}_{\text{2}}$O. The exception is that the attention score is quantized into a binary value, with 1 indicating above average and 0 indicating below average. \\
\textbf{Last Token Attention Score~(LTAS)}~~TOVA~\cite{tova} uses last token's attention score as importance indicator.

\subsection{Eviction Scope Construction}
\label{sec:esc}
Due to the auto-regressive nature of LLMs, recent tokens in the cache participate in less attention computation than earlier tokens. Therefore, their recorded importance scores for some attention-based methods can be underestimated and thus get wrongly evicted. To this end, an eviction scope should be constructed to carefully select tokens allowed to be evicted.

The dominant mean of constructing eviction scope is 
 \textit{\textbf{local window}}, which assumes that tokens outside of a 
local window of size $r$ have accumulated sufficient information on their importance.

\subsection{Preliminary Experiments}
\label{sec:preliminary}
In our controlled preliminary experiments, we are interested in how different importance 
score calculation methods behave in terms of consistency with respect to 
their full KV cache version. After that, we also explore another way of constructing 
eviction scope in addition to local window.

\begin{figure}[t]
	\centering
	\scalebox{0.415}{\includegraphics{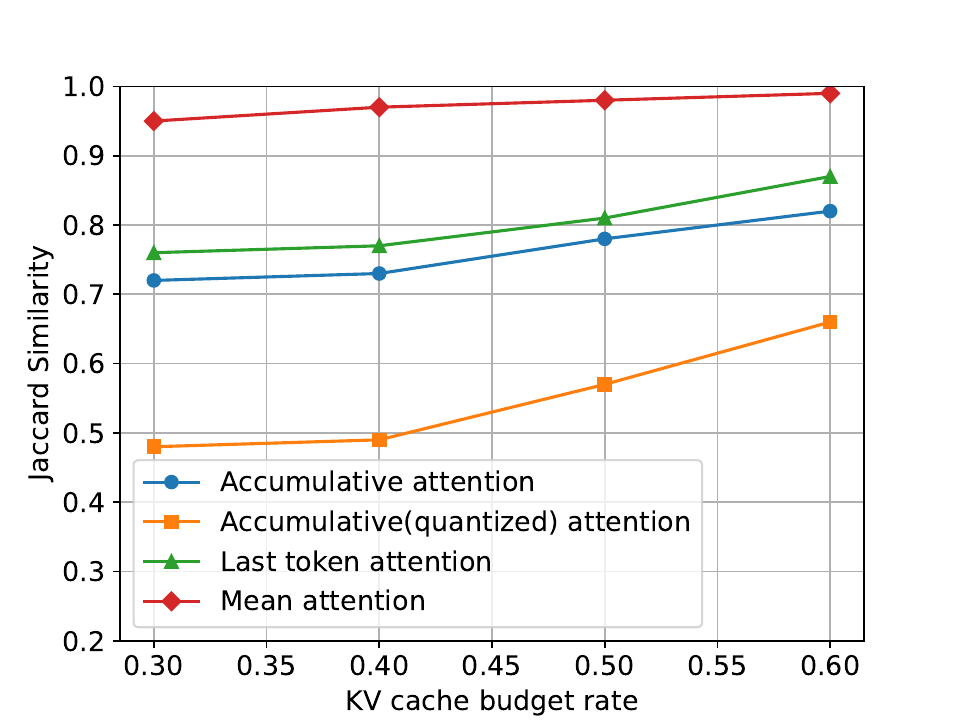}}
	\caption{Consistency of different importance calculation methods w.r.t their full KV cache variant.}
	\label{fig:consistency}
\end{figure}

\paragraph{Setup} We examine the Jaccard similarity between the top-$B$ important tokens 
derived by various importance score calculation methods and those derived when a full 
KV cache is available. The higher the similarity, the more effectively the importance 
calculation method harnesses local information to approximate the global one. We evaluate all 
attention-based methods~(i.e., AAS, AQAS, and LTAS) listed in \secref{sec:isc} and set 
the local window size $r$ to 0. More specifically, we use LLaMa2-7B-Chat\footnote{We also 
conduct experiments upon other LLMs like WizardLM-7B, and similar results are observed.} as the LLM and take the 805 instructions from AlpacaEval~\cite{alpaca_eval} as prompt, generating a response for each instruction via greedy decoding. For KV cache budget from $\{0.3,0.4,0.5,0.6\}$, we compute the Jaccard similarity at each token position, averaged over sequence length, attention heads, and layers. 

\paragraph{Results} The results are shown in \figref{fig:consistency}. AAS shows a clear advantage over AQAS across all budget rates, indicating the importance of a full-precision attention score when the relative importance of tokens cannot be distinguished via binary value. LTAS has higher consistency than AAS, which we attribute to the fact that 
LTAS does not suffer from the recency bias that AAS and AQAS exhibit due to the accumulation operation. However, LTAS computes importance score based on a single token, which might bear high variability. Based on the above results, we advocate the use of \textit{Mean Attention Score~(MAS) to gauge the importance of each token}. MAS divides each token's accumulative attention score by how many times that token is attended by future tokens. As shown in \figref{fig:consistency}, MAS has remarkably higher consistency among all methods, achieving over 0.9 Jaccard similarity even at 0.3 cache budget rate. This verifies that MAS effectively alleviates recency bias and can better retain high-influential tokens.
\begin{figure}[t]
	\centering
	\scalebox{0.48}{\includegraphics{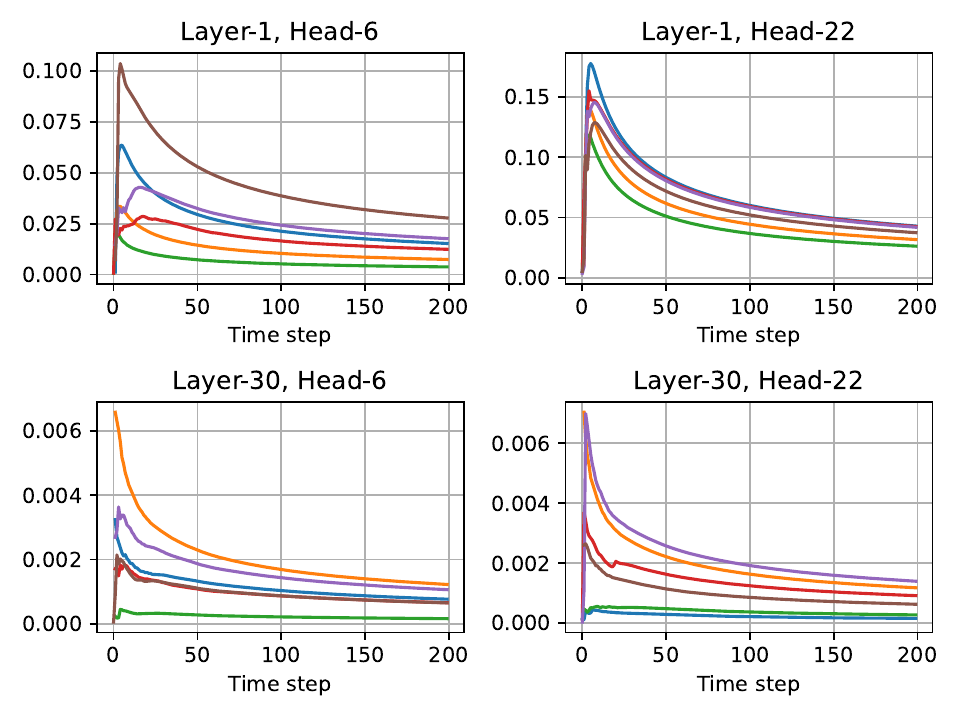}}
    \caption{Illustration of persistence of attention robustness. We extract attention scores and compute the standard deviation from LLaMa2-7B-Chat.}
	\label{fig:std}
\end{figure}

\paragraph{Is Local Window the Optimal Way to Construct Eviction Scope?} 
The local window approach has been widely used in conjunction with attention-based importance calculation methods to prevent recent tokens from being evicted. The underlying rationale is that the accumulated attention score is not sufficiently indicative until a specified threshold, i.e., the window size $r$ is reached. 

Based on the commendable consistency of MAS discussed in the previous paragraph, here we propose another way to construct the eviction scope which exploits a phenomenon termed \textit{persistence of attention robustness} that we find ubiquitously exists in large language models.  
It states that the standard deviation of the attention probabilities a token receives 
from future tokens typically undergoes a brief ascending phase before settling into a stable 
decline, regardless of the model layer and attention head. \figref{fig:std} clearly illustrates the phenomenon. We also observe that the ascending phase of a non-trivial portion of tokens only takes a relatively small number of steps, i.e., $\leq 50$, suggesting it might be sub-optimal to only consider tokens at least $r$ steps away for eviction scope construction. 

To this end, we propose a new way to construct the eviction scope utilizing the \textit{standard deviation} of attention scores. Concretely, we maintain another $B$-sized array for each attention head, keep track of the accumulative squared attention score, and compute the standard deviation of each in-cache token $x$ using $\text{Std}(x)=\sqrt{\frac{\text{Acc}^{\text{square}}(x)}{\text{Count}(x)}-(\frac{\text{Acc}(x)}{\text{Count}(x)})^2}$. In practice, Acc, Acc$^{\text{sqaure}}$, and Count are all $B$-sized tensor and the standard deviation of all in-cache tokens are computed in parallel. Then, instead of the most recent $r$ tokens, we exclude tokens having top-$r$ standard deviation from eviction scope and remove the key-value vectors corresponding to the token with the lowest mean attention score. 

\begin{figure}[t]
	\centering
	\scalebox{0.385}{\includegraphics{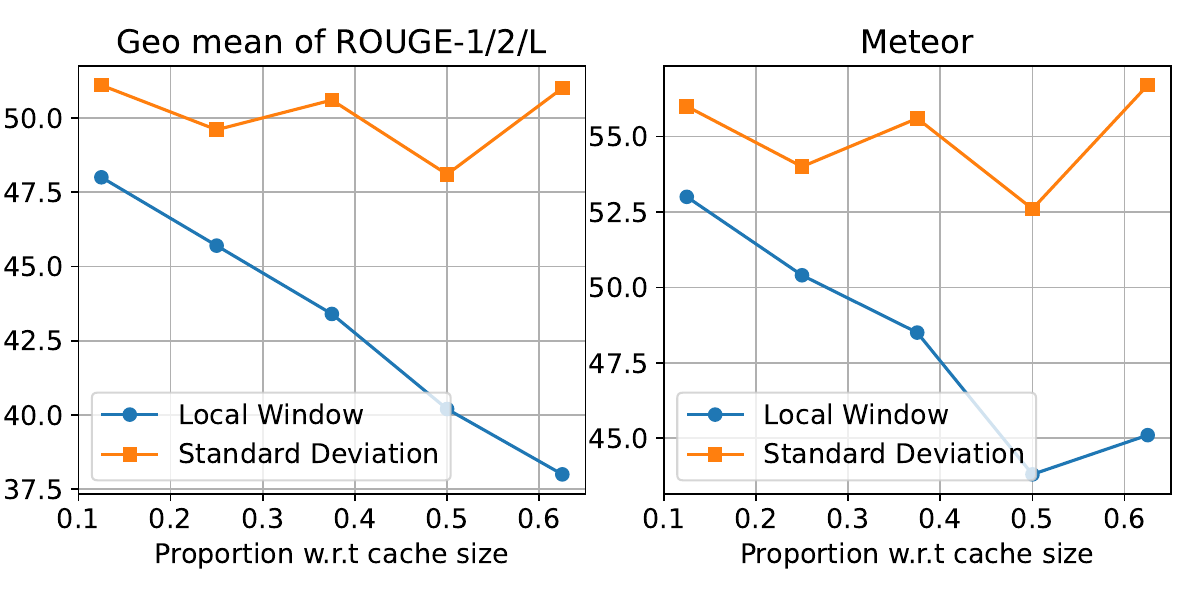}}
    \caption{Results of MAS paired with local window and standard deviation on text summarization.}
	\label{fig:scope}
\end{figure}
We validate the effectiveness of both eviction scopes on a news summarization task with LLaMa2-7B-Chat and CNN/Daily Mail~\cite{cnndm} dataset. Since summarization is a typical long-input-short-output task, we only perform KV eviction at the prefilling stage with a 0.5 compression rate and compare the output against the full KV cache version. \figref{fig:scope} shows the geometric mean of ROUGE-1/2/L~\cite{rouge} and METEOR~\cite{meteor} for eviction scopes of different sizes. Standard deviation yields outputs with considerably higher quality while being less sensitive to the size of the eviction scope.

\paragraph{RoCo} Combining mean attention score for importance score calculation and standard deviation for eviction scope construction, we introduce RoCo as a \underline{R}\underline{o}bust \underline{C}ache \underline{o}mission policy for key-value constrained generative LLM inference.

\section{Comprehensive Evaluation}
The goal of the eviction policy is to control the memory usage of key-value cache under a fixed budget while retaining the generation quality of LLMs as much as possible. 
In this section, we perform an empirical evaluation of the effectiveness of various eviction policies by taking the generated output with a full KV cache as the reference and comparing KV-restricted generations against it.
\subsection{Experiment Setup}
We describe the experimental setup used throughout our evaluation, including evaluation tasks, metrics, datasets, and compared eviction policies.
\subsubsection{Tasks and Metrics}
\label{sec:tasks}
To broadly cover real-world use cases, we evaluate using four different types of tasks: language modeling, abstractive text summarization, original context reconstruction, and instruction following.
\paragraph{Language Modeling} Language modeling task assesses the ability of LLMs to predict the next token given the preceding context. In the key-value-constrained scenario, a successful eviction policy should be able to detect and remove KV cache of unimportant tokens. Following prior works~\cite{llminfinite,xiao2023efficient,tova}, we adopt perplexity as the evaluation metric.
\paragraph{Abstractive Text Summarization} Abstractive summarization requires extracting the most salient information provided in the input and generating a concise summary for it. Since the summary is usually much shorter compared to the input, we only perform cache eviction during the prefilling stage. We report BLEU~\cite{bleu}, ROUGE~\cite{rouge}, and METEOR~\cite{meteor} scores as the evaluation metrics.
\paragraph{Original Context Reconstruction} Given the constrained incomplete key-value cache of an input document, the task of original context reconstruction measures how well the limited KV cache retains the essential information from the original context. BLEU and ROUGE scores are used as evaluation metrics.
\paragraph{Instruction Following} Instruction following~\cite{flan} requires an LLM to generate a proper response for a given user instruction. We apply KV cache eviction at the auto-regressive decoding stage since the model output tends to be more verbose. In addition to BLEU and ROUGE scores, we also opt for a pairwise comparison paradigm to evaluate the generated responses against those generated by text-davincci-003.
\subsubsection{Datasets} 
We use the following datasets as the testbed for tasks described in \secref{sec:tasks}.
\paragraph{OpenWebText} OpenWebText is an open-source replication of the WebText dataset from OpenAI. We randomly sample 200 documents to form the test set for the language modeling task.
\paragraph{XSum} Xsum~\cite{xsum} comprises BBC articles from the years 2010 to 2017, encompassing a broad spectrum of topics.
\paragraph{CNN/Daily Mail} CNN/Daily Mail~\cite{cnndm} contains articles from the CNN and the Daily Mail newspapers, representing a different distribution from XSum. We use this dataset for both summarization and original context reconstruction.
\paragraph{AlpacaEval} AlpacaEval~\cite{alpaca_eval} is a model-based automatic evaluation benchmark for instruction-following LLMs. It comprises 805 instructions spanning a diverse range of scenarios.
\subsubsection{Models}
Following prior works~\cite{h2o,tova}, we employ LLaMa2-7B-base for language modeling and LLaMa2-7B-Chat for the remaining tasks. We also include WizardLM-7B~\cite{wizardlm} as another strong instruction-tuned LLM for tasks except for language modeling.

\subsubsection{Compared Eviction Policies}
\begin{table}[h]
    \small
    \centering
    \begin{tabular}{l|cc}
     \toprule
     & \multicolumn{1}{l}{\textbf{Importance Score}} & \multicolumn{1}{l}{\textbf{Eviction Scope}} \\
     \midrule
    Random       & -    & -                \\
    StreamLLM   & -    & local window    \\
    ScissorHands & AQAS & local window       \\
    H$_{\text{2}}$O          & AAS  & local window       \\
    TOVA         & LTAS & -                \\
    RoCo         & MAS  & standard deviation \\
    \bottomrule
    \end{tabular}
    \caption{Eviction policies considered in this paper. The definition of importance score and eviction scope are introduced in \secref{sec:isc} and \secref{sec:esc}, respectively.}
    \label{tab:policies}
\end{table}
\begin{table*}[t]
    \centering
    \small
    \begin{tabular}{cc|ccccc|ccccc}
    \toprule
    \multicolumn{1}{c}{\multirow{2}{*}{\textbf{Models}}} & \multirow{2}{*}{\textbf{Methods}} & \multicolumn{5}{c}{\textbf{XSum}}         & \multicolumn{5}{c}{\textbf{CNN/DM}}       \\
    \multicolumn{1}{c}{} &              & BLEU & Meteor & R-1  & R-2  & R-L  & BLEU & Meteor & R-1  & R-2  & R-L  \\
    \midrule
    \multirow{6}{*}{LLaMa2-7B$_{\text{chat}}$}             & Random                   & 16.7 & 35.1 & 41.8 & 22.8 & 33.3 & 15.4 & 30.8 & 39.0 & 19.9 & 26.2 \\
                         & StreamLLM   & 11.9 & 35.5   & 42.7 & 17.6 & 31.3 & 19.2 & 40.0   & 49.5 & 24.3 & 30.7 \\
                         & ScissorHands & 29.3 & 50.4   & 56.5 & 35.6 & 46.5 & 27.8 & 46.4   & 57.7 & 33.4 & 40.7 \\
                         & H$_{\text{2}}$O          & 37.3 & 55.8   & 62.2 & 44.2 & 54.2 & 31.1 & 47.6   & 58.9 & 36.8 & 43.7 \\
                         & TOVA         & 20.5 & 42.4   & 48.9 & 26.9 & 39.9 & 21.5 & 41.7   & 53.1 & 27.4 & 34.8 \\
                         & RoCo         & \textbf{43.4} & \textbf{60.5}   & \textbf{65.0} & \textbf{48.6} & \textbf{57.8} & \textbf{33.2} & \textbf{49.3}   & \textbf{61.1} & \textbf{39.2} & \textbf{46.0} \\
    \midrule
    \multirow{6}{*}{WizardLM-7B}             & Random                   &12.6  &30.0  &36.3  &19.1  &28.0  &12.4  &27.8  &39.6  &17.0  &23.9  \\
                         & StreamLLM   &7.1  &27.3    &36.2  &13.4  &25.6  &8.8  &25.9    &39.0  &14.4  &23.8  \\
                         & ScissorHands &29.8  &48.6    &57.2  &38.5  &47.7  &24.6  &41.2    &54.3  &29.1  &35.7  \\
                         & H$_{\text{2}}$O          &30.5  &50.2    &57.6  &40.8  &49.7  &27.7  &43.3    &56.4  &32.5  &39.2  \\
                         & TOVA         &15.0  &36.2    &46.1  &23.9  &34.8  &14.5  &32.7    &45.9  &19.4  &27.9  \\
                         & RoCo         &\textbf{35.7}  &\textbf{56.6}    &\textbf{61.4}  &\textbf{45.5}  &\textbf{53.7}  &\textbf{30.2}  &\textbf{45.9}    &\textbf{59.1}  &\textbf{35.6}  &\textbf{42.4}  \\
    \bottomrule
    \end{tabular}
    \caption{Performance of different eviction policies on abstractive text summarization tasks at 0.5 KV cache rate.}
    \label{table:sum}
\end{table*}
We consider the following baseline eviction policies, with their importance score calculation methods and eviction scope listed in \tabref{tab:policies}:
\begin{itemize}[itemsep=1pt,parsep=2pt,topsep=1pt]
    \item Random: evicting a randomly selected key-value pair from the cache.
    \item StreamLLM~\cite{xiao2023efficient}: evicting the key-value pair corresponding to the first token after 4 initial attention sink tokens.
    \item ScissorHands~\cite{liu2023scissorhands}: evicting the key-value pair corresponding to the token with the smallest accumulative quantized attention score outside the local window of size $r$.
    \item H$_{\text{2}}$O~\cite{h2o}: evicting the key-value pair corresponding to the token with the smallest accumulative attention score outside of the local window of size $r$.
    \item TOVA~\cite{tova}: evicting the key-value pair corresponding to the token with the smallest last token attention score.
\end{itemize}
\subsubsection{Other Details}
Our implementation is based on Pytorch~\cite{pytorch} and HuggingFace Transformers~\cite{wolf-etal-2020-transformers}. To improve the stability of outputs produced by LLMs, we employ greedy decoding for all generative tasks. For prefilling stage eviction, we set budget size $B$ by multiplying input token length with a compression rate~(e.g., 0.5) and directly specify $B$ as some integer for decoding stage eviction because the output length is unknown. The size of the local window is set to half of the KV cache budget following H$_{\text{2}}$O, i.e. $r=B/2$.
\begin{figure}[t]
	\centering
	\scalebox{0.41}{\includegraphics{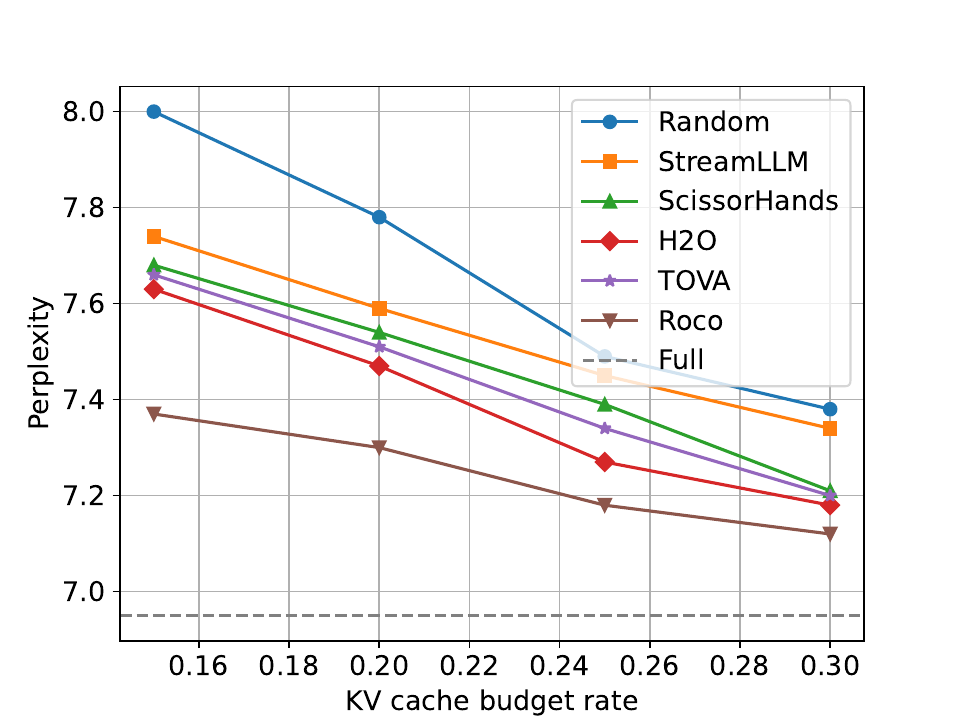}}
    \caption{Performance of different eviction policies on language modeling task based on LLaMa2-7B.}
	\label{fig:ppl}
\end{figure}
\begin{table}[t]
    \centering
    \small
    \begin{tabular}{cc|ccc}
    \toprule
    \multirow{2}{*}{\textbf{Models}}    & \multirow{2}{*}{\textbf{Methods}} & \multicolumn{3}{c}{\textbf{CNN/DM}} \\
                               &                          & BLEU   & R-1  & R-L  \\
    \midrule
    \multirow{6}{*}{LLaMa2-7B$_{\text{chat}}$} & Random                   &4.9             &26.7      &18.1      \\
                               & StreamLLM                &15.1             &37.6      &28.5      \\
                               & ScissorHands             &16.9             &45.9      &33.7      \\
                               & H$_{\text{2}}$O                      &20.8             &50.6      &40.7      \\
                               & TOVA                     &11.2             & 43.1     &30.3      \\
                               & RoCo                     & \textbf{28.1}            &\textbf{57.7}      &\textbf{49.2}      \\
    \midrule
    \multirow{6}{*}{WizardLM-7B} & Random                   &6.3             &30.3      &19.6      \\
                               & StreamLLM                &5.3             &24.7      &16.3      \\
                               & ScissorHands             &17.4             &48.3      &33.0      \\
                               & H$_{\text{2}}$O                      &20.8             &50.8      &38.5      \\
                               & TOVA                     &12.5             &41.7      &27.7      \\
                               & RoCo                     &\textbf{29.0}             &\textbf{58.9}      &\textbf{47.5}      \\
    \bottomrule
    \end{tabular}
    \caption{Performance of different eviction policies on context reconstruction task at 0.5 KV cache rate.}
    \label{table:rec}
\end{table}
\begin{table*}[t]
    \centering
    \small
    \begin{tabular}{cc|cccc}
    \toprule
    \multicolumn{1}{c}{\multirow{2}{*}{\textbf{Models}}} & \multirow{2}{*}{\textbf{Methods}} & \multicolumn{4}{c}{\textbf{AlpacaEval}}               \\
    \multicolumn{1}{c}{} &              & BLEU & ROUGE-1  & ROUGE-2  & ROUGE-L  \\
    \midrule
    \multirow{6}{*}{LLaMa2-7B$_{\text{chat}}$}             & Random                   &45.6  &66.6  &49.7  &56.1     \\
                         & StreamLLM   &47.3  &66.6    &51.1  &57.1    \\
                         & ScissorHands &62.1  &76.5    &63.9  &68.8     \\
                         & H$_{\text{2}}$O          &63.0  &77.5   &65.7  &69.9      \\
                         & TOVA         &60.1 &75.3    &63.8  &68.3    \\
                         & RoCo         &\textbf{66.3}  &\textbf{79.7}    &\textbf{68.1}  &\textbf{72.5}     \\
    \midrule
    \multirow{6}{*}{WizardLM-7B}             & Random                   &40.6  &62.9  &45.6  &52.8     \\
                         & StreamLLM   &43.0  &63.9    &47.8  &54.5     \\
                         & ScissorHands &57.2  &73.3    &60.9  &65.3     \\
                         & H$_{\text{2}}$O          &58.6  &74.1    &62.1  &66.8     \\
                         & TOVA         &59.6  &75.0    &63.1  &68.0     \\
                         & RoCo         &\textbf{62.2}  &\textbf{76.4}    &\textbf{64.5}  &\textbf{69.5}    \\
    \bottomrule
    \end{tabular}
    \caption{Performance of different eviction policies on AlpacaEval at 250-token KV cache budget.}
    \label{table:alpacaeval}
\end{table*}
\subsection{Main Results}
\begin{table}[t]
    \centering
    \small
    \begin{tabular}{cccc}
    \toprule
    \textbf{Budget} & \textbf{StreamLLM} & \textbf{H$_{\text{2}}$O} & \textbf{RoCo} \\
    \midrule
    200              & 72.0(-3.6)               & 74.5(-1.1)         & 75.2(-0.4)          \\
    250              & 72.9(-2.7)               & 74.8(-0.8)         & 75.5(-0.1)          \\
    \bottomrule
    \end{tabular}
    \caption{AlpacaEval win rates against text-davincci-003 judged by GPT-4. Numbers in the parenthesis denote the performance drop compared to full KV cache~(500 token output length on average with 75.6 win rate).}
    \label{table:gpt4}
\end{table}
\paragraph{Overview} We report the results of RoCo alongside compared baseline eviction policies on language modeling, text summarization, original context reconstruction, and instruction following tasks in \figref{fig:ppl}, \tabref{table:sum}, \tabref{table:alpacaeval}, and \tabref{table:rec}. 
It can be seen that RoCo consistently outperforms previous methods by significant margins across all tasks and models. The advantage of Roco is particularly more evident at a low KV cache budget rate. As demonstrated in \figref{fig:ppl}, the gap between Roco and the second best method H$_{\text{2}}$O reaches 0.2 at 0.15 budget rate. We also notice that Roco yields larger improvements on generative inference tasks compared to language modeling. This is because, in generative tasks, future token predictions are dependent on previous model generations, where minor errors can accumulate and lead to large divergence. 
on the AlpacaEval benchmark, RoCo not only delivers the best overall performance in conventional metrics like BLEU and ROUGE, but also achieves comparable win rates as judged by GPT-4~(\tabref{table:gpt4}). In contrast, H$_{\text{2}}$O shows notable quality declines, and the gap is even larger for StreamLLM. Overall, the experiment results validate the effectiveness of Roco in terms of retaining model performance during key-value-constrained inference.
\paragraph{Attention Matters for KV Eviction} As the only two policies that do not utilize attention-related information, Random and StreamLLM show significantly inferior performance compared to attention-based methods on all tasks. This observation aligns with the role of key-value vectors in attention computation, where the attention score serves natural indicator of token importance.
\subsection{Discussion}
\paragraph{Extension to Grouped-Query Attention~(GQA)} GQA, along with its extreme case Multi-Query Attention~(MQA) has gained increasing adoption in powerful LLMs like Mistral~\cite{mistral}. We extend the attention-based eviction policy to GQA and MQA by taking the group-wise averaged attention score and using it to update the importance score according to \secref{sec:isc}. The results of Zephyr-7B~\cite{tunstall2023zephyr} in Appendix \ref{sec:appendix_a} validate the effectiveness of our extension.
\paragraph{Overhead} An ideal eviction policy should avoid introducing much extra overhead since LLMs are already memory and computational-intensive. The memory overhead induced by Roco is $L\times H\times B\times 3$, which is negligible given the reduced KV cache footprint $L\times 2\times H\times (S-B)\times d^\prime$, where $S$ is the non-evicted full sequence length. Moreover, different from the auto-regressive decoding stage which is I/O-bounded, the prefilling stage is computation-bounded. However, prior policies usually evict one token every time the cache is full and encode the next token, turning the prefilling stage into I/O-bounded. To accelerate key-value constrained prompt encoding, RoCo allows for performing evict-and-encode in a block-wise manner. The block size $b$ controls the number of tokens being freed and encoded within one eviction step. We examine the effect of block-wise eviction using LLaMa2-7B-Chat on XSum summarization task. \tabref{table:blockwise} shows that such block-wise eviction greatly speeds up prefilling while retaining similar output quality as token-wise eviction. More results are deferred to Appendix \ref{sec:appendix_b} due to space limit.
\begin{table}[t]
    \centering
    \small
    \begin{tabular}{cccc}
    \toprule
    \textbf{Block Size} & \textbf{BLEU} & \textbf{ROUGE-2} & \textbf{Speed up} \\
    \midrule
    1                   & 41.2          & 46.2             & 1.0x              \\
    2                   & 41.1          & 46.3             & 2.0x              \\
    4                   & 41.1          & 46.2             & 4.0x              \\
    8                   & 40.4          & 45.5             & 8.0x              \\
    16                  & 40.2          & 45.4             & 16.0x             \\
    \bottomrule
    \end{tabular}
    \caption{Summarization results of block-wise eviction using RoCo. }
    \label{table:blockwise}
\end{table}
\paragraph{Integrated Package} Finally, we open-source EasyKV, a software package dedicated to key-value constrained inference accompanying this research. It is designed to be fully compatible with existing LLMs with different attention variants and enables flexible configuration of diverse eviction policies, cache budgets, and application scenarios.

\section{Conclusion}
This paper studies key-value restricted language model inference. To shed light on the effectiveness of existing eviction policies, we conduct comprehensive comparative analysis by decomposing eviction policy into importance score calculation and eviction scope construction. We identify the inconsistency and instability of prior policie and introduce RoCo, a robust cache omission policy with improved downstream performance. We also release EasyKV, the accompanying library for versatile key-value constrained LLM inference.

% \clearpage
% \newpage
\bibliography{custom}

\begin{thebibliography}{46}
\expandafter\ifx\csname natexlab\endcsname\relax\def\natexlab#1{#1}\fi

\bibitem[{Ainslie et~al.(2023)Ainslie, Lee-Thorp, de~Jong, Zemlyanskiy, Lebron,
  and Sanghai}]{gqa}
Joshua Ainslie, James Lee-Thorp, Michiel de~Jong, Yury Zemlyanskiy, Federico
  Lebron, and Sumit Sanghai. 2023.
\newblock \href {https://doi.org/10.18653/v1/2023.emnlp-main.298} {{GQA}:
  Training generalized multi-query transformer models from multi-head
  checkpoints}.
\newblock In \emph{Proceedings of the 2023 Conference on Empirical Methods in
  Natural Language Processing}, pages 4895--4901, Singapore. Association for
  Computational Linguistics.

\bibitem[{Ainslie et~al.(2020)Ainslie, Ontanon, Alberti, Cvicek, Fisher, Pham,
  Ravula, Sanghai, Wang, and Yang}]{etc}
Joshua Ainslie, Santiago Ontanon, Chris Alberti, Vaclav Cvicek, Zachary Fisher,
  Philip Pham, Anirudh Ravula, Sumit Sanghai, Qifan Wang, and Li~Yang. 2020.
\newblock Etc: Encoding long and structured inputs in transformers.
\newblock \emph{arXiv preprint arXiv:2004.08483}.

\bibitem[{An et~al.(2023)An, Gong, Zhong, Li, Zhang, Kong, and Qiu}]{leval}
Chenxin An, Shansan Gong, Ming Zhong, Mukai Li, Jun Zhang, Lingpeng Kong, and
  Xipeng Qiu. 2023.
\newblock L-eval: Instituting standardized evaluation for long context language
  models.
\newblock \emph{arXiv preprint arXiv:2307.11088}.

\bibitem[{Ashkboos et~al.(2024)Ashkboos, Croci, Nascimento, Hoefler, and
  Hensman}]{slicegpt}
Saleh Ashkboos, Maximilian~L Croci, Marcelo Gennari~do Nascimento, Torsten
  Hoefler, and James Hensman. 2024.
\newblock Slicegpt: Compress large language models by deleting rows and
  columns.
\newblock \emph{arXiv preprint arXiv:2401.15024}.

\bibitem[{Bai et~al.(2023)Bai, Lv, Zhang, Lyu, Tang, Huang, Du, Liu, Zeng, Hou
  et~al.}]{longbench}
Yushi Bai, Xin Lv, Jiajie Zhang, Hongchang Lyu, Jiankai Tang, Zhidian Huang,
  Zhengxiao Du, Xiao Liu, Aohan Zeng, Lei Hou, et~al. 2023.
\newblock Longbench: A bilingual, multitask benchmark for long context
  understanding.
\newblock \emph{arXiv preprint arXiv:2308.14508}.

\bibitem[{Banerjee and Lavie(2005)}]{meteor}
Satanjeev Banerjee and Alon Lavie. 2005.
\newblock \href {https://aclanthology.org/W05-0909} {{METEOR}: An automatic
  metric for {MT} evaluation with improved correlation with human judgments}.
\newblock In \emph{Proceedings of the {ACL} Workshop on Intrinsic and Extrinsic
  Evaluation Measures for Machine Translation and/or Summarization}, pages
  65--72, Ann Arbor, Michigan. Association for Computational Linguistics.

\bibitem[{Beltagy et~al.(2020)Beltagy, Peters, and Cohan}]{longformer}
Iz~Beltagy, Matthew~E Peters, and Arman Cohan. 2020.
\newblock Longformer: The long-document transformer.
\newblock \emph{arXiv preprint arXiv:2004.05150}.

\bibitem[{Child et~al.(2019)Child, Gray, Radford, and
  Sutskever}]{child2019generating}
Rewon Child, Scott Gray, Alec Radford, and Ilya Sutskever. 2019.
\newblock Generating long sequences with sparse transformers.
\newblock \emph{arXiv preprint arXiv:1904.10509}.

\bibitem[{Choromanski et~al.(2020)Choromanski, Likhosherstov, Dohan, Song,
  Gane, Sarlos, Hawkins, Davis, Mohiuddin, Kaiser et~al.}]{performer}
Krzysztof Choromanski, Valerii Likhosherstov, David Dohan, Xingyou Song,
  Andreea Gane, Tamas Sarlos, Peter Hawkins, Jared Davis, Afroz Mohiuddin,
  Lukasz Kaiser, et~al. 2020.
\newblock Rethinking attention with performers.
\newblock \emph{arXiv preprint arXiv:2009.14794}.

\bibitem[{Dao(2023)}]{dao2023flashattention}
Tri Dao. 2023.
\newblock Flashattention-2: Faster attention with better parallelism and work
  partitioning.
\newblock \emph{arXiv preprint arXiv:2307.08691}.

\bibitem[{Dao et~al.(2022{\natexlab{a}})Dao, Chen, Sohoni, Desai, Poli, Grogan,
  Liu, Rao, Rudra, and R{\'e}}]{dao2022monarch}
Tri Dao, Beidi Chen, Nimit~S Sohoni, Arjun Desai, Michael Poli, Jessica Grogan,
  Alexander Liu, Aniruddh Rao, Atri Rudra, and Christopher R{\'e}.
  2022{\natexlab{a}}.
\newblock Monarch: Expressive structured matrices for efficient and accurate
  training.
\newblock In \emph{International Conference on Machine Learning}, pages
  4690--4721. PMLR.

\bibitem[{Dao et~al.(2022{\natexlab{b}})Dao, Fu, Ermon, Rudra, and
  R{\'e}}]{dao2022flashattention}
Tri Dao, Dan Fu, Stefano Ermon, Atri Rudra, and Christopher R{\'e}.
  2022{\natexlab{b}}.
\newblock Flashattention: Fast and memory-efficient exact attention with
  io-awareness.
\newblock \emph{Advances in Neural Information Processing Systems},
  35:16344--16359.

\bibitem[{Dettmers et~al.(2022)Dettmers, Lewis, Belkada, and
  Zettlemoyer}]{dettmers2022llm}
Tim Dettmers, Mike Lewis, Younes Belkada, and Luke Zettlemoyer. 2022.
\newblock Llm. int8 (): 8-bit matrix multiplication for transformers at scale.
\newblock \emph{arXiv preprint arXiv:2208.07339}.

\bibitem[{Frantar and Alistarh(2023)}]{frantar2023massive}
Elias Frantar and Dan Alistarh. 2023.
\newblock Massive language models can be accurately pruned in one-shot.
\newblock \emph{arXiv preprint arXiv:2301.00774}.

\bibitem[{Frantar et~al.(2022)Frantar, Ashkboos, Hoefler, and Alistarh}]{gptq}
Elias Frantar, Saleh Ashkboos, Torsten Hoefler, and Dan Alistarh. 2022.
\newblock Gptq: Accurate post-training quantization for generative pre-trained
  transformers.
\newblock \emph{arXiv preprint arXiv:2210.17323}.

\bibitem[{Han et~al.(2023)Han, Wang, Xiong, Chen, Ji, and Wang}]{llminfinite}
Chi Han, Qifan Wang, Wenhan Xiong, Yu~Chen, Heng Ji, and Sinong Wang. 2023.
\newblock Lm-infinite: Simple on-the-fly length generalization for large
  language models.
\newblock \emph{arXiv preprint arXiv:2308.16137}.

\bibitem[{Jiang et~al.(2023)Jiang, Sablayrolles, Mensch, Bamford, Chaplot,
  Casas, Bressand, Lengyel, Lample, Saulnier et~al.}]{mistral}
Albert~Q Jiang, Alexandre Sablayrolles, Arthur Mensch, Chris Bamford,
  Devendra~Singh Chaplot, Diego de~las Casas, Florian Bressand, Gianna Lengyel,
  Guillaume Lample, Lucile Saulnier, et~al. 2023.
\newblock Mistral 7b.
\newblock \emph{arXiv preprint arXiv:2310.06825}.

\bibitem[{Li et~al.(2023)Li, Zhang, Dubois, Taori, Gulrajani, Guestrin, Liang,
  and Hashimoto}]{alpaca_eval}
Xuechen Li, Tianyi Zhang, Yann Dubois, Rohan Taori, Ishaan Gulrajani, Carlos
  Guestrin, Percy Liang, and Tatsunori~B. Hashimoto. 2023.
\newblock Alpacaeval: An automatic evaluator of instruction-following models.
\newblock \url{https://github.com/tatsu-lab/alpaca_eval}.

\bibitem[{Lin(2004)}]{rouge}
Chin-Yew Lin. 2004.
\newblock \href {https://aclanthology.org/W04-1013} {{ROUGE}: A package for
  automatic evaluation of summaries}.
\newblock In \emph{Text Summarization Branches Out}, pages 74--81, Barcelona,
  Spain. Association for Computational Linguistics.

\bibitem[{Liu et~al.(2023{\natexlab{a}})Liu, Lin, Hewitt, Paranjape,
  Bevilacqua, Petroni, and Liang}]{liu2023lost}
Nelson~F Liu, Kevin Lin, John Hewitt, Ashwin Paranjape, Michele Bevilacqua,
  Fabio Petroni, and Percy Liang. 2023{\natexlab{a}}.
\newblock Lost in the middle: How language models use long contexts.
\newblock \emph{arXiv preprint arXiv:2307.03172}.

\bibitem[{Liu et~al.(2023{\natexlab{b}})Liu, Desai, Liao, Wang, Xie, Xu,
  Kyrillidis, and Shrivastava}]{liu2023scissorhands}
Zichang Liu, Aditya Desai, Fangshuo Liao, Weitao Wang, Victor Xie, Zhaozhuo Xu,
  Anastasios Kyrillidis, and Anshumali Shrivastava. 2023{\natexlab{b}}.
\newblock Scissorhands: Exploiting the persistence of importance hypothesis for
  llm kv cache compression at test time.
\newblock \emph{arXiv preprint arXiv:2305.17118}.

\bibitem[{Narayan et~al.(2018)Narayan, Cohen, and Lapata}]{xsum}
Shashi Narayan, Shay~B. Cohen, and Mirella Lapata. 2018.
\newblock Don't give me the details, just the summary! topic-aware
  convolutional neural networks for extreme summarization.
\newblock \emph{ArXiv}, abs/1808.08745.

\bibitem[{Oren et~al.(2024)Oren, Hassid, Adi, and Schwartz}]{tova}
Matanel Oren, Michael Hassid, Yossi Adi, and Roy Schwartz. 2024.
\newblock Transformers are multi-state rnns.
\newblock \emph{arXiv preprint arXiv:2401.06104}.

\bibitem[{Papineni et~al.(2002)Papineni, Roukos, Ward, and Zhu}]{bleu}
Kishore Papineni, Salim Roukos, Todd Ward, and Wei-Jing Zhu. 2002.
\newblock \href {https://doi.org/10.3115/1073083.1073135} {{B}leu: a method for
  automatic evaluation of machine translation}.
\newblock In \emph{Proceedings of the 40th Annual Meeting of the Association
  for Computational Linguistics}, pages 311--318, Philadelphia, Pennsylvania,
  USA. Association for Computational Linguistics.

\bibitem[{Paszke et~al.(2019)Paszke, Gross, Massa, Lerer, Bradbury, Chanan,
  Killeen, Lin, Gimelshein, Antiga et~al.}]{pytorch}
Adam Paszke, Sam Gross, Francisco Massa, Adam Lerer, James Bradbury, Gregory
  Chanan, Trevor Killeen, Zeming Lin, Natalia Gimelshein, Luca Antiga, et~al.
  2019.
\newblock Pytorch: An imperative style, high-performance deep learning library.
\newblock \emph{Advances in neural information processing systems}, 32.

\bibitem[{Qin et~al.(2022)Qin, Han, Sun, Li, Kong, Barnes, and
  Zhong}]{qin-etal-2022-devil}
Zhen Qin, Xiaodong Han, Weixuan Sun, Dongxu Li, Lingpeng Kong, Nick Barnes, and
  Yiran Zhong. 2022.
\newblock \href {https://doi.org/10.18653/v1/2022.emnlp-main.473} {The devil in
  linear transformer}.
\newblock In \emph{Proceedings of the 2022 Conference on Empirical Methods in
  Natural Language Processing}, pages 7025--7041, Abu Dhabi, United Arab
  Emirates. Association for Computational Linguistics.

\bibitem[{Radford et~al.(2019)Radford, Wu, Child, Luan, Amodei, Sutskever
  et~al.}]{radford2019language}
Alec Radford, Jeffrey Wu, Rewon Child, David Luan, Dario Amodei, Ilya
  Sutskever, et~al. 2019.
\newblock Language models are unsupervised multitask learners.
\newblock \emph{OpenAI blog}, 1(8):9.

\bibitem[{See et~al.(2017)See, Liu, and Manning}]{cnndm}
Abigail See, Peter~J. Liu, and Christopher~D. Manning. 2017.
\newblock \href {https://doi.org/10.18653/v1/P17-1099} {Get to the point:
  Summarization with pointer-generator networks}.
\newblock In \emph{Proceedings of the 55th Annual Meeting of the Association
  for Computational Linguistics (Volume 1: Long Papers)}, pages 1073--1083,
  Vancouver, Canada. Association for Computational Linguistics.

\bibitem[{Shazeer(2019)}]{mqa}
Noam Shazeer. 2019.
\newblock Fast transformer decoding: One write-head is all you need.
\newblock \emph{arXiv preprint arXiv:1911.02150}.

\bibitem[{Sun et~al.(2023)Sun, Yin, Li, Wu, Qiu, and Kong}]{sun2023corex}
Qiushi Sun, Zhangyue Yin, Xiang Li, Zhiyong Wu, Xipeng Qiu, and Lingpeng Kong.
  2023.
\newblock Corex: Pushing the boundaries of complex reasoning through
  multi-model collaboration.
\newblock \emph{arXiv preprint arXiv:2310.00280}.

\bibitem[{Thoppilan et~al.(2022)Thoppilan, De~Freitas, Hall, Shazeer,
  Kulshreshtha, Cheng, Jin, Bos, Baker, Du et~al.}]{thoppilan2022lamda}
Romal Thoppilan, Daniel De~Freitas, Jamie Hall, Noam Shazeer, Apoorv
  Kulshreshtha, Heng-Tze Cheng, Alicia Jin, Taylor Bos, Leslie Baker, Yu~Du,
  et~al. 2022.
\newblock Lamda: Language models for dialog applications.
\newblock \emph{arXiv preprint arXiv:2201.08239}.

\bibitem[{Touvron et~al.(2023)Touvron, Martin, Stone, Albert, Almahairi,
  Babaei, Bashlykov, Batra, Bhargava, Bhosale et~al.}]{llama2}
Hugo Touvron, Louis Martin, Kevin Stone, Peter Albert, Amjad Almahairi, Yasmine
  Babaei, Nikolay Bashlykov, Soumya Batra, Prajjwal Bhargava, Shruti Bhosale,
  et~al. 2023.
\newblock Llama 2: Open foundation and fine-tuned chat models.
\newblock \emph{arXiv preprint arXiv:2307.09288}.

\bibitem[{Tunstall et~al.(2023)Tunstall, Beeching, Lambert, Rajani, Rasul,
  Belkada, Huang, von Werra, Fourrier, Habib et~al.}]{tunstall2023zephyr}
Lewis Tunstall, Edward Beeching, Nathan Lambert, Nazneen Rajani, Kashif Rasul,
  Younes Belkada, Shengyi Huang, Leandro von Werra, Cl{\'e}mentine Fourrier,
  Nathan Habib, et~al. 2023.
\newblock Zephyr: Direct distillation of lm alignment.
\newblock \emph{arXiv preprint arXiv:2310.16944}.

\bibitem[{Vaswani et~al.(2017)Vaswani, Shazeer, Parmar, Uszkoreit, Jones,
  Gomez, Kaiser, and Polosukhin}]{vaswani2017attention}
Ashish Vaswani, Noam Shazeer, Niki Parmar, Jakob Uszkoreit, Llion Jones,
  Aidan~N Gomez, {\L}ukasz Kaiser, and Illia Polosukhin. 2017.
\newblock Attention is all you need.
\newblock \emph{Advances in neural information processing systems}, 30.

\bibitem[{Wang et~al.(2020)Wang, Li, Khabsa, Fang, and Ma}]{linformer}
Sinong Wang, Belinda~Z Li, Madian Khabsa, Han Fang, and Hao Ma. 2020.
\newblock Linformer: Self-attention with linear complexity.
\newblock \emph{arXiv preprint arXiv:2006.04768}.

\bibitem[{Wang et~al.(2023)Wang, Kordi, Mishra, Liu, Smith, Khashabi, and
  Hajishirzi}]{wang-etal-2023-self-instruct}
Yizhong Wang, Yeganeh Kordi, Swaroop Mishra, Alisa Liu, Noah~A. Smith, Daniel
  Khashabi, and Hannaneh Hajishirzi. 2023.
\newblock \href {https://doi.org/10.18653/v1/2023.acl-long.754} {Self-instruct:
  Aligning language models with self-generated instructions}.
\newblock In \emph{Proceedings of the 61st Annual Meeting of the Association
  for Computational Linguistics (Volume 1: Long Papers)}, pages 13484--13508,
  Toronto, Canada. Association for Computational Linguistics.

\bibitem[{Wei et~al.(2021)Wei, Bosma, Zhao, Guu, Yu, Lester, Du, Dai, and
  Le}]{flan}
Jason Wei, Maarten Bosma, Vincent~Y Zhao, Kelvin Guu, Adams~Wei Yu, Brian
  Lester, Nan Du, Andrew~M Dai, and Quoc~V Le. 2021.
\newblock Finetuned language models are zero-shot learners.
\newblock \emph{arXiv preprint arXiv:2109.01652}.

\bibitem[{Wei et~al.(2022)Wei, Tay, Bommasani, Raffel, Zoph, Borgeaud,
  Yogatama, Bosma, Zhou, Metzler et~al.}]{wei2022emergent}
Jason Wei, Yi~Tay, Rishi Bommasani, Colin Raffel, Barret Zoph, Sebastian
  Borgeaud, Dani Yogatama, Maarten Bosma, Denny Zhou, Donald Metzler, et~al.
  2022.
\newblock Emergent abilities of large language models.
\newblock \emph{arXiv preprint arXiv:2206.07682}.

\bibitem[{Wolf et~al.(2020)Wolf, Debut, Sanh, Chaumond, Delangue, Moi, Cistac,
  Rault, Louf, Funtowicz, Davison, Shleifer, von Platen, Ma, Jernite, Plu, Xu,
  Scao, Gugger, Drame, Lhoest, and Rush}]{wolf-etal-2020-transformers}
Thomas Wolf, Lysandre Debut, Victor Sanh, Julien Chaumond, Clement Delangue,
  Anthony Moi, Pierric Cistac, Tim Rault, Rémi Louf, Morgan Funtowicz, Joe
  Davison, Sam Shleifer, Patrick von Platen, Clara Ma, Yacine Jernite, Julien
  Plu, Canwen Xu, Teven~Le Scao, Sylvain Gugger, Mariama Drame, Quentin Lhoest,
  and Alexander~M. Rush. 2020.
\newblock \href {https://www.aclweb.org/anthology/2020.emnlp-demos.6}
  {Transformers: State-of-the-art natural language processing}.
\newblock In \emph{Proceedings of the 2020 Conference on Empirical Methods in
  Natural Language Processing: System Demonstrations}, pages 38--45, Online.
  Association for Computational Linguistics.

\bibitem[{Wu et~al.(2024)Wu, Han, Ding, Weng, Liu, Yao, Yu, and
  Kong}]{wu2024copilot}
Zhiyong Wu, Chengcheng Han, Zichen Ding, Zhenmin Weng, Zhoumianze Liu, Shunyu
  Yao, Tao Yu, and Lingpeng Kong. 2024.
\newblock Os-copilot: Towards generalist computer agents with self-improvement.
\newblock \emph{arXiv preprint arXiv:2402.07456}.

\bibitem[{Xia et~al.(2023)Xia, Gao, Zeng, and Chen}]{xia2023sheared}
Mengzhou Xia, Tianyu Gao, Zhiyuan Zeng, and Danqi Chen. 2023.
\newblock Sheared llama: Accelerating language model pre-training via
  structured pruning.
\newblock \emph{arXiv preprint arXiv:2310.06694}.

\bibitem[{Xiao et~al.(2023{\natexlab{a}})Xiao, Lin, Seznec, Wu, Demouth, and
  Han}]{smoothquant}
Guangxuan Xiao, Ji~Lin, Mickael Seznec, Hao Wu, Julien Demouth, and Song Han.
  2023{\natexlab{a}}.
\newblock Smoothquant: Accurate and efficient post-training quantization for
  large language models.
\newblock In \emph{International Conference on Machine Learning}, pages
  38087--38099. PMLR.

\bibitem[{Xiao et~al.(2023{\natexlab{b}})Xiao, Tian, Chen, Han, and
  Lewis}]{xiao2023efficient}
Guangxuan Xiao, Yuandong Tian, Beidi Chen, Song Han, and Mike Lewis.
  2023{\natexlab{b}}.
\newblock Efficient streaming language models with attention sinks.
\newblock \emph{arXiv preprint arXiv:2309.17453}.

\bibitem[{Xu et~al.(2023)Xu, Sun, Zheng, Geng, Zhao, Feng, Tao, and
  Jiang}]{wizardlm}
Can Xu, Qingfeng Sun, Kai Zheng, Xiubo Geng, Pu~Zhao, Jiazhan Feng, Chongyang
  Tao, and Daxin Jiang. 2023.
\newblock Wizardlm: Empowering large language models to follow complex
  instructions.
\newblock \emph{arXiv preprint arXiv:2304.12244}.

\bibitem[{Zaheer et~al.(2020)Zaheer, Guruganesh, Dubey, Ainslie, Alberti,
  Ontanon, Pham, Ravula, Wang, Yang et~al.}]{bigbird}
Manzil Zaheer, Guru Guruganesh, Kumar~Avinava Dubey, Joshua Ainslie, Chris
  Alberti, Santiago Ontanon, Philip Pham, Anirudh Ravula, Qifan Wang, Li~Yang,
  et~al. 2020.
\newblock Big bird: Transformers for longer sequences.
\newblock \emph{Advances in neural information processing systems},
  33:17283--17297.

\bibitem[{Zhang et~al.(2023)Zhang, Sheng, Zhou, Chen, Zheng, Cai, Song, Tian,
  R{\'e}, Barrett et~al.}]{h2o}
Zhenyu Zhang, Ying Sheng, Tianyi Zhou, Tianlong Chen, Lianmin Zheng, Ruisi Cai,
  Zhao Song, Yuandong Tian, Christopher R{\'e}, Clark Barrett, et~al. 2023.
\newblock H $ \_2 $ o: Heavy-hitter oracle for efficient generative inference
  of large language models.
\newblock \emph{arXiv preprint arXiv:2306.14048}.

\end{thebibliography}
\bibliographystyle{acl_natbib}
\clearpage
\newpage
\appendix
\section{More Experimental Results}
\paragraph{Prompt Template} For instruction-tuned/aligned LLMs used in the main experiments, we strictly follow their system prompt used during training. Specifically, we list the prompt template used for LLaMa2-7B/13B-Chat, WizardLM-7B, and Zephyr-7B as follows:
\begin{itemize}
    \item LLaMa2-7B/13B-Chat: [INST] <<SYS>> You are a helpful, respectful and honest assistant. Always answer as helpfully as possible, while being safe.  Your answers should not include any harmful, unethical, racist, sexist, toxic, dangerous, or illegal content. Please ensure that your responses are socially unbiased and positive in nature. If a question does not make any sense, or is not factually coherent, explain why instead of answering something not correct. If you don't know the answer to a question, please don't share false information. <</SYS>> \{instruction\}[/INST]
    \item WizardLM-7B: Below is an instruction that describes a task. Write a response that appropriately completes the request. Instruction: \{instruction\} Response:
    \item Zephyr-7B: <|system|> You are a friendly chatbot who always responds in a helpful and detailed manner to the user's questions.</s> <|user|> \{instruction\}</s> <|assistant|>
\end{itemize}
\label{sec:appendix_a}
\paragraph{Extension to GQA/MQA}
We extend existing attention-based eviction policies into GQA and MQA by taking the group-wise averaged attention scores and using it to update the importance score. To verify our extension, we conduct experiments using Zephyr-7B~\cite{tunstall2023zephyr}, an instruction-tuned and aligned version of Mistral-7B~\cite{mistral} on the text summarization task. Specifically, Zephyr-7B employs GQA and has 8 key-value heads and 32 query heads, rendering a 4x replication for each key-value vector pair.

The results are shown in \tabref{table:zephyr}. We can see that, attention-based eviction policies still exhibit better performance compared to Random and StreamLLM, showing that the group-wise average operation can effectively reflect the importance of each token across all query heads within its group.
\paragraph{Results on Larger LLMs} In addition to 7B-scale LLMs, we also examine the effectiveness of RoCo alongside other eviction policies on 13B-scale LLMs. To this end, we conduct a text summarization experiment using LLaMa2-13B-Chat~\cite{llama2} and report the results in \tabref{table:llama213b}. We observe that, given the same KV cache budget, LLaMa2-13B-Chat achieves higher BLEU and ROUGE scores than LLaMa2-7B-Chat. It indicates that a larger model dimension may contain more redundancy in some less informative intermediate activations. This observation is inspiring because it implies that we can preserve more performance when using more powerful LLMs.

\begin{table*}[t!]
    \centering
    \small
    \begin{tabular}{cc|ccccc|ccccc}
    \toprule
    \multicolumn{1}{c}{\multirow{2}{*}{\textbf{Models}}} & \multirow{2}{*}{\textbf{Methods}} & \multicolumn{5}{c}{\textbf{XSum}}         & \multicolumn{5}{c}{\textbf{CNN/DM}}       \\
    \multicolumn{1}{c}{} &              & BLEU & Meteor & R-1  & R-2  & R-L  & BLEU & Meteor & R-1  & R-2  & R-L  \\
    \midrule
    \multirow{6}{*}{Zephyr-7B} & Random &17.0  &39.0  &48.6  &25.2  &34.4  &22.8  &43.2  &54.5  &29.0  &36.1  \\
                         & StreamLLM   &12.0  &35.0    &45.1 &19.3  &29.0  &20.2  &41.2    &51.8  &26.1  &31.9  \\
                         & ScissorHands &26.6  &48.7    &57.4  &34.3  &44.2  &29.4  &49.1    &60.4  &36.0  &44.0  \\
                         & H$_{\text{2}}$O   &29.6  &49.9    &58.9  &38.3  &47.2  &34.9  &52.3    &63.6  &41.3  &48.4  \\
                         & TOVA         &16.8  &40.7    &50.5  &24.6  &34.9  &25.5  &45.8    &57.5  &32.2  &38.6  \\
                         & RoCo         &\textbf{33.6}  &\textbf{54.9}    &\textbf{62.6}  &\textbf{42.4}  &\textbf{50.4}  &\textbf{36.6}  &\textbf{53.8}    &\textbf{64.6}  &\textbf{42.6}  &\textbf{50.0}  \\
    \bottomrule
    \end{tabular}
    \caption{Performance of Zephyr-7B using different eviction policies on abstractive text summarization tasks at 0.5 KV cache rate.}
    \label{table:zephyr}
\end{table*}
\begin{table*}[t!]
    \centering
    \small
    \begin{tabular}{cc|ccccc|ccccc}
    \toprule
    \multicolumn{1}{c}{\multirow{2}{*}{\textbf{Models}}} & \multirow{2}{*}{\textbf{Methods}} & \multicolumn{5}{c}{\textbf{XSum}}         & \multicolumn{5}{c}{\textbf{CNN/DM}}       \\
    \multicolumn{1}{c}{} &              & BLEU & Meteor & R-1  & R-2  & R-L  & BLEU & Meteor & R-1  & R-2  & R-L  \\
    \midrule
    \multirow{4}{*}{LLaMa2-13B$_{\text{Chat}}$} & Random &14.9  &27.7  &30.0  &17.9  &24.5  &9.0  &18.8  &22.3  &11.7  &15.1  \\
                         & StreamLLM   &14.7  &35.7    &41.7 &19.8  &31.2  &26.0  &43.1    &53.2  &30.4  &36.6  \\
                         & H$_{\text{2}}$O   &38.3  &55.9    &61.0  &44.5  &53.4  &38.4  &52.5    &62.3  &42.6  &48.6  \\
                         & RoCo         &\textbf{45.8}  &\textbf{62.2}    &\textbf{66.6}  &\textbf{51.5}  &\textbf{59.3}  &\textbf{41.4}  &\textbf{54.1}    &\textbf{65.1}  &\textbf{47.0}  &\textbf{53.0}  \\
    \bottomrule
    \end{tabular}
    \caption{Performance of LLaMa2-13B-Chat using different eviction policies on abstractive text summarization tasks at 0.5 KV cache rate.}
    \label{table:llama213b}
\end{table*}
\begin{table}[h]
    \centering
    \small
    \begin{tabular}{cccc}
    \toprule
    \textbf{Block Size} & \textbf{BLEU} & \textbf{ROUGE-2} & \textbf{Speed up} \\
    \midrule
    1                   & 26.5          & 35.4            & 1.0x              \\
    2                   & 26.6          & 35.4             & 2.0x              \\
    4                   & 26.5          & 35.4             & 4.0x              \\
    8                   & 26.9          & 35.8             & 8.0x              \\
    16                  & 26.4          & 35.4             & 16.0x             \\
    \bottomrule
    \end{tabular}
    \caption{Performance of Zephyr-7B on text summarization using RoCo. Larger block size only leads to a slight performance decline while significantly speed-up the prefilling stage.}
    \label{table:block}
\end{table}

\section{More Results on Block-wise Eviction} 
\label{sec:appendix_b}
To accelerate key-value constrained prompt encoding, we extend the per-token evict-and-encode scheme to a block-wise manner. We report results on text summarization with various block sizes using Zephyr-7B in \tabref{table:block}. With a larger block size, more tokens are evicted with less reliable importance scores, thereby resulting in some influential tokens being wrongly evicted. Nevertheless, the performance drop is tolerable given the significant speedup of prompt encoding, especially when confronted with long-context tasks~\cite{liu2023lost,longbench,leval}.
\section{Case Study} 
\label{sec:appendix_c}
\begin{figure*}[t!]
	\centering
	\scalebox{0.54}{\includegraphics{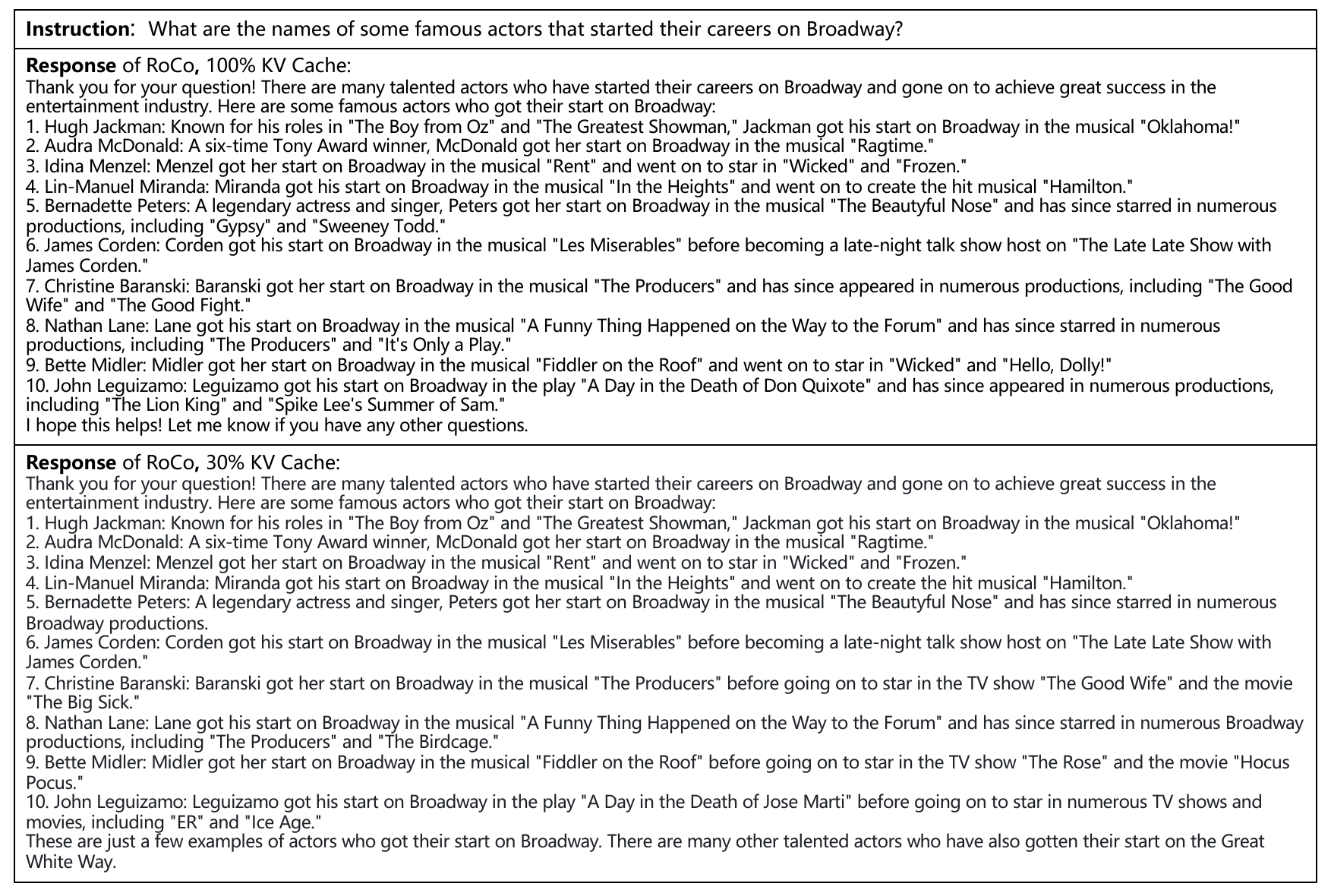}}
    \caption{Case study of LLaMa2-7B-Chat generated response given a specific instruction. The response generated with 30\% KV cache using RoCo retains almost all content in the original response.}
	\label{fig:case_study}
\end{figure*}

To obtain a straightforward impression on the generation quality when RoCo is applied for key-value constrained inference, we present responses generated by LLaMa2-7B-Chat given the instruction ``\textit{What are the names of some famous actors that started their careers on Broadway?}''. The responses at different KV cache budget are shown in \figref{fig:case_study}. At 0.3 KV cache budget rate, RoCo generates a response containing the same actor/actress names as the one conditioned on a full KV cache, demonstrating the commendable ability of RoCo to selectively retain useful key-value states and maintain coherent generation.

\end{document}